\documentclass[11pt,a4paper]{article}
\usepackage[hyperref]{acl2020}
\usepackage{times}
\usepackage{latexsym}

\usepackage{url}
\usepackage[T1]{fontenc}
\usepackage[draft]{todonotes}
\usepackage{amsmath}
\usepackage{amsfonts}
\usepackage{graphicx}
\usepackage{multirow, tabu}
\usepackage[title]{appendix}
\usepackage{float}
\usepackage{siunitx,tabularx}

\definecolor{cadmiumgreen}{rgb}{0.0, 0.42, 0.24}
\definecolor{bostonuniversityred}{rgb}{0.8, 0.0, 0.0}

\usepackage{microtype}

\usepackage{cleveref}
\crefformat{section}{\S#2#1#3} 
\crefformat{subsection}{\S#2#1#3}
\crefformat{subsubsection}{\S#2#1#3}

\aclfinalcopy 

\title{A Generative Model for Joint \\Natural Language Understanding and Generation}

\author{Bo-Hsiang Tseng$^{1}$\thanks{$^{*}$Work done while the author was an intern at Apple. }, Jianpeng Cheng${}^2$, Yimai Fang${}^2$ and David Vandyke${}^2$ \\
${}^1$Engineering Department, University of Cambridge, UK \\
${}^2$Apple \\
\texttt{bht26@cam.ac.uk} \\
\texttt{\{jianpeng.cheng, yimai\_fang, dvandyke\}@apple.com}\\
}

\date{}

\begin{document}
\maketitle
\begin{abstract}
Natural language understanding (NLU) and natural language generation (NLG) are two fundamental and related tasks in building task-oriented dialogue systems with opposite objectives: NLU tackles the transformation from natural language to formal representations, whereas NLG does the reverse. A key to success in either task is parallel training data which is expensive to obtain at a large scale.
In this work, we propose a generative model which couples NLU and NLG through a shared latent variable. This approach allows us to explore both spaces of natural language and formal representations, and facilitates information sharing through the latent space to eventually benefit NLU and NLG.
Our model achieves state-of-the-art performance on two dialogue datasets with both flat and tree-structured formal representations.
We also show that the model can be trained in a semi-supervised fashion by utilising unlabelled data to boost its performance.

\end{abstract}

\section{Introduction}
Natural language understanding (NLU) and natural language generation (NLG) are two fundamental tasks in building task-oriented dialogue systems. In a modern dialogue system, an NLU module first converts a user utterance, provided by an automatic speech recognition model, into a formal representation. The representation is then consumed by a downstream dialogue state tracker to update a belief state which represents an aggregated user goal. Based on the current belief state, a policy network decides the formal representation of the system response.  This is finally used by an NLG module to generate the system response\cite{YOUNG2010150}. 

It can be observed that NLU and NLG have opposite goals: NLU aims to map natural language to formal representations, while NLG generates utterances from their semantics. In research literature, NLU and NLG are well-studied as separate problems. State-of-the-art NLU systems tackle the task as classification \cite{zhang2016joint} or as structured prediction or generation \cite{damonte2019practical}, depending on the formal representations  which can be flat slot-value pairs \cite{henderson2014word}, first-order logical form \cite{zettlemoyer2012learning}, or structured queries \cite{yu2018typesql,pasupat-etal-2019-span}. On the other hand, approaches to NLG vary from pipelined approach subsuming content planning and surface realisation \cite{stent2004trainable} to more recent end-to-end sequence generation \cite{wen2015semantically, duvsek2020evaluating}.

\begin{figure}[t!]
  \centering
  \includegraphics[width=\linewidth]{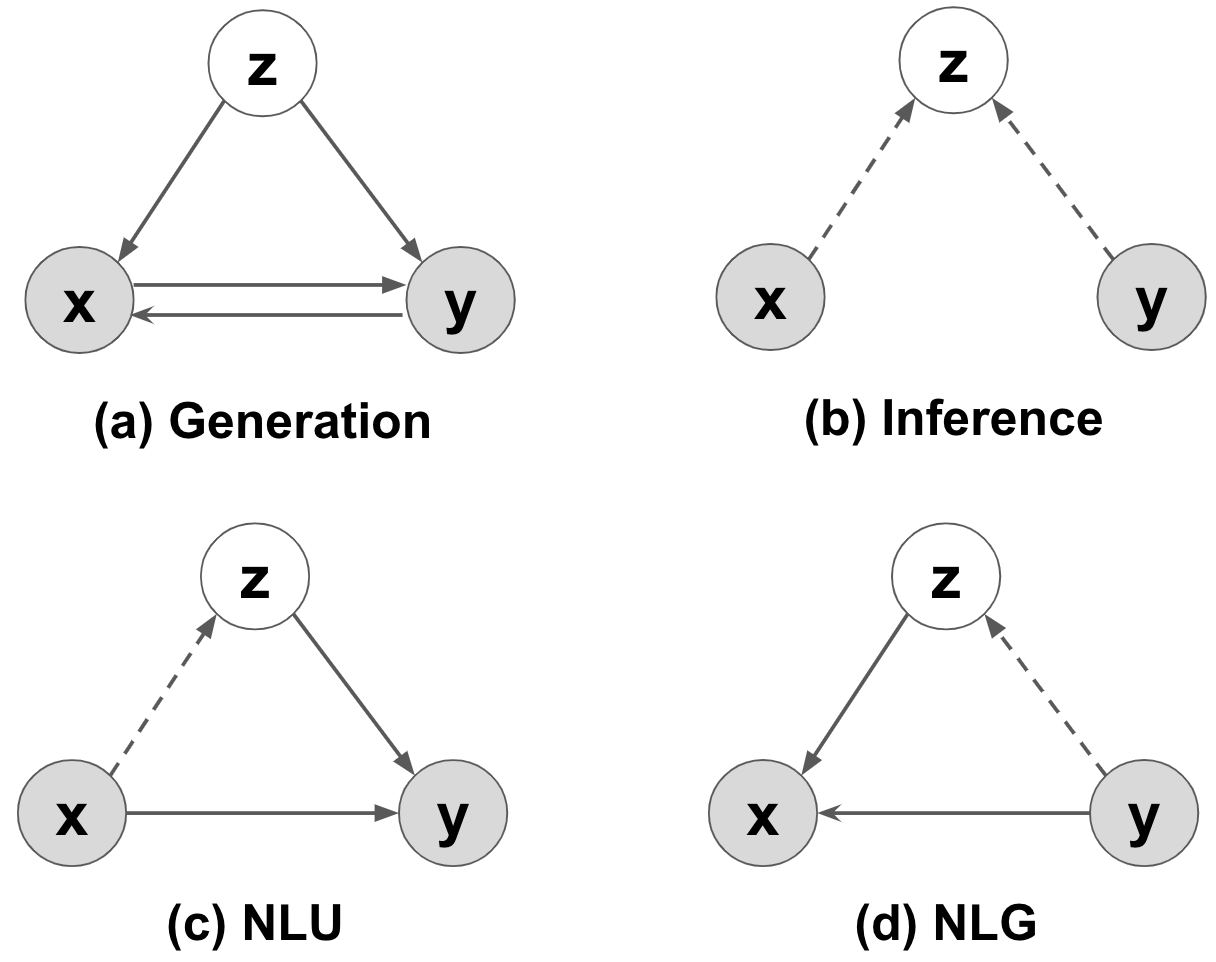}
  \caption{Generation and inference process in our model, and how NLU and NLG are achieved. $x$ and $y$ denotes utterances and formal representations respectively; $z$ represents the shared latent variable for $x$ and $y$.}
  \label{fig:model}
\end{figure}

However, the duality between NLU and NLG has been less explored. In fact, both tasks can be treated as a translation problem: NLU converts natural language to formal language while NLG does the reverse. Both tasks require a substantial amount of utterance and representation pairs to succeed, and such data is costly to collect due to the complexity of annotation involved. Although unannotated data for either natural language or formal representations can be easily obtained, it is less clear how they can be leveraged as the two languages stand in different space.

In this paper, we propose a generative model for \textbf{J}oint natural language \textbf{U}nderstanding and \textbf{G}eneration (\textbf{JUG}), which couples NLU and NLG with a latent variable representing the shared intent between natural language and formal representations.
We aim to learn the association between two discrete spaces through a continuous latent variable which facilitates information sharing between two tasks. 
Moreover, JUG can be trained in a semi-supervised fashion, which enables us to explore each space of natural language and formal representations when unlabelled data is accessible. 
We examine our model on two dialogue datasets with different formal representations: the E2E dataset \cite{novikova-etal-2017-e2e} where the semantics are represented as a collection of slot-value pairs; and a more recent weather dataset \cite{balakrishnan2019constrained} where the formal representations are tree-structured. Experimental results show that our model improves over standalone NLU/NLG models and existing methods on both tasks; and the performance can be further boosted by utilising unlabelled data.

\section{Model}
Our key assumption is that there exists an abstract latent variable $z$ underlying a pair of utterance $x$ and formal representation $y$.
In our generative model, this abstract intent guides the standard conditional generation of either NLG or NLU (Figure 1a).
Meanwhile, $z$ can be inferred from either utterance $x$, or formal representation $y$ (Figure 1b).
That means performing NLU requires us to infer the $z$ from $x$, after which the formal representation $y$ is generated conditioning on both $z$ and $x$ (Figure 1c), and vice-versa for NLG (Figure 1d).
In the following, we will explain the model details, starting with NLG.

\subsection{NLG}
\label{sec:nlu-nlg}
As mentioned above, the task of NLG requires us to infer $z$ from $y$, and then generate $x$ using both $z$ and $y$.
We choose the posterior distribution $q(z|y)$ to be Gaussian.
The task of inferring $z$ can then be recast to computing mean $\mu$ and standard deviation $\sigma$ of the Gaussian distribution using an NLG encoder.
To do this, we use a bi-directional LSTM \cite{hochreiter1997long} to encode formal representation $y$.
which is linearised and represented as a sequence of symbols.  After encoding, we obtain a list of hidden vectors $\mathbf{H}$, with each representing the concatenation of forward and backward LSTM states.
These hidden vectors are then average-pooled and passed through two feed-forward neural networks to compute mean $\pmb{\mu}_{y,z}$ and standard deviation $\pmb{\sigma}_{y,z}$ vectors of the posterior $q(z|y)$.
\begin{align}
\begin{split}
\label{eq:nlg-enc}
    & \mathbf{H} = \text{Bi-LSTM}(\mathbf{y}) \\
    & \mathbf{\mathbf{\bar{h}}} = \text{Pooling}(\mathbf{H}) \\
    & \pmb{\mu}_{y,z} = \mathbf{W}_{\mu}\mathbf{\bar{h}} + \mathbf{b}_{\mu} \\
    & \pmb{\sigma}_{y,z} = \mathbf{W}_{\sigma}\mathbf{\bar{h}} + \mathbf{b}_{\sigma}
\end{split}
\end{align}   
where $\mathbf{W}$ and $\mathbf{b}$ represent neural network weights and bias.
Then the latent vector $\mathbf{z}$ can be sampled from the approximated posterior using the re-parameterisation trick of \citet{kingma2013auto}:
\begin{align}
\label{eq:nlg-sample}
\begin{split}
 & \pmb{\epsilon} \sim \mathcal{N}(0, \mathbf{I}) \\
 & \mathbf{z} = \pmb{\mu}_{y,z} + \pmb{\sigma}_{y, z} \pmb{\epsilon}
\end{split}
\end{align}   
The final step is to generate natural language $x$ based on latent variable $z$ and formal representation $y$. We use an LSTM decoder relying on both $z$ and $y$ via attention mechanism \cite{bahdanau2014neural}. At each time step, the decoder computes:
\begin{align}
\label{eq:nlg-dec}
\begin{split}
    & \mathbf{g}_{i}^{x} = \text{LSTM}(\mathbf{g}_{i-1}^{x}, \mathbf{x}_{i-1}) \\
    & \mathbf{c}_{i} = \text{attention}(\mathbf{g}_{i}^{x}, \mathbf{H}) \\
    & p(x_{i}) = \text{softmax}( \mathbf{W_v} [\mathbf{c}_{i} \hspace{-0.5mm} \oplus \hspace{-0.5mm}  \mathbf{g}_{i}^{x} \hspace{-0.5mm} \oplus \hspace{-0.5mm}    \mathbf{z}]+\mathbf{b}_v)
\end{split}
\end{align}
where $\oplus$ denotes concatenation. $\mathbf{x}_{i-1}$ is the word vector of input token; $\mathbf{g}_{i}^{x}$ is the corresponding decoder hidden state and $p(x_{i})$ is the output token distribution at time step $i$.


\subsection{NLU}
NLU performs the reverse procedures of NLG.
First, an NLU encoder infers the latent variable $z$ from utterance $x$. The encoder uses a bi-directional LSTM to convert
the utterance into a list of hidden states. These hidden states are pooled and passed through 
feed-forward neural networks to compute the mean $\pmb{\mu}_{x,z}$ and standard deviation $\pmb{\sigma}_{x,z}$ of the posterior $q(z|x)$. 
This procedure follows Equation \ref{eq:nlg-enc} in NLG.

However, note that a subtle difference between natural language and formal language is that the former is  ambiguous while the later is precisely defined. 
This makes NLU a many-to-one mapping problem but NLG is one-to-many. 
To better reflect the fact that the NLU output requires less variance, 
when decoding we choose the latent vector $\mathbf{z}$ in NLU to be the mean vector $\pmb{\mu}_{x,z}$, instead of sampling it from $q(z|x)$ like Equation \ref{eq:nlg-sample}.\footnote{Note that it is still necessary to compute the standard deviation $\pmb{\sigma}_{x,z}$ in NLU, since the term is needed for optimisation. See more details in Section \ref{sec:optimisation}.}  

After the latent vector is obtained, 
the formal representation $y$ is predicted from both $z$ and $x$ using an NLU decoder. Since the space of $y$ depends on the formal language construct, we consider two common scenarios in dialogue systems.
In the first scenario, $y$ is represented as a set of slot-value pairs, e.g., \{\emph{food type}=\emph{British}, \emph{area}=\emph{north}\} in restaurant search domain \cite{mrkvsic2017neural}. The decoder here consists of several classifiers, one for each slot, to predict the corresponding values.\footnote{Each slot has a set of corresponding values plus a special one \texttt{not\_mention}.} Each classifier is modelled by a 1-layer feed-forward neural network that takes $z$ as input:
\begin{align}
\label{eq:nlu-dec1}
\begin{split}
   & p(y_{s}) = \text{softmax}(\mathbf{W_s} \mathbf{z} + \mathbf{b_s}) \\
\end{split}
\end{align}
where $p(y_{s})$ is the predicted value distribution of slot $s$.

In the second scenario, $y$ is a tree-structured formal representation \cite{banarescu-etal-2013-abstract}.
We then generate $y$ as a linearised token sequence using an LSTM decoder relying on both $z$ and $x$ via the standard attention mechanism \cite{bahdanau2014neural}. The decoding procedure follows exactly Equation \ref{eq:nlg-dec}.

\subsection{Model Summary} 
One flexibility of the JUG model comes from the fact that it has two ways to infer the shared latent variable $z$ through either $x$ or $y$; and the inferred $z$ can aid the generation of both $x$ and $y$. 
In this next section, we show how this shared latent variable enables the JUG model to explore unlabelled $x$ and $y$,
while aligning the learned meanings inside the latent space.

\section{Optimisation}
\label{sec:optimisation}
We now describe how JUG can be optimised with a pair of $x$ and $y$ (\S\ref{sec:p_xy}), and also unpaired  $x$ or $y$ (\S\ref{sec:p_x}). 
We specifically discuss the prior choice of JUG objectives in \S\ref{choice of p}.
A combined objective can be thus derived for semi-supervised learning: a practical scenario when we have a small set of labelled data but abundant unlabelled ones (\S\ref{opt summary}).

\subsection{Optimising $p(x,y)$}
\label{sec:p_xy}
Given a pair of utterance $x$ and formal representation $y$, our objective is to maximise the log-likelihood of the joint probability $p(x,y)$:
\begin{equation}
\log p(x,y) = \log \int_z p(x, y, z)
\end{equation}
The optimisation task is not directly tractable since it requires us to marginalise out the latent variable $z$. However, it can be solved by following the standard practice of neural variational inference \cite{kingma2013auto}. An objective based on the variational lower bound can be derived as
\begin{equation}
\begin{split}
    \mathcal{L}_{x,y} & = \mathbb{E}_{q(z|x)} \log p(y|z, x) + \mathbb{E}_{q(z|x)} \log p(x|z, y) \\
     & \,\,\,\,\, -  \text{KL} [q(z|x) || p(z)]
\end{split}
\label{eq:xy_x}
\end{equation}
where the first term on the right side is the NLU model; the second term is the reconstruction of $x$; and the last term denotes the Kullback$-$Leibler divergence between the approximate posterior $q(z|x)$ with the prior $p(z)$. We defer the discussion of prior to Section \ref{choice of p} and detailed derivations to Appendix.

The symmetry between utterance and semantics offers an alternative way of inferring the posterior through the approximation $q(z|y)$.
Analogously we can derive a variational optimisation objective:
\begin{equation}
\label{eq:xy_y}
\begin{split}
    \mathcal{L}_{y,x} & = \mathbb{E}_{q(z|y)} \log p(x|z, y) + \mathbb{E}_{q(z|y)} \log p(y|z, x) \\
     & \,\,\,\,\, -  \text{KL} [q(z|y) || p(z)]
\end{split}
\end{equation}
where the first term is the NLG model; the second term is the reconstruction of $y$; and the last term denotes the KL divergence.

It can be observed that our model has two posterior inference paths from either $x$ or $y$, and also two generation paths. All paths can be optimised.

\subsection{Optimising $p(x)$ or $p(y)$}
\label{sec:p_x}
Additionally, when we have access to unlabelled utterance $x$ (or formal representation $y$), the optimisation objective of JUG is the marginal likelihood  $p(x)$ (or $p(y)$):
\begin{equation}
\log p(x) = \log \int_y \int_z p(x, y, z)
\end{equation}
Note that both $z$ and $y$ are unobserved in this case.

We can develop an objective based on the variational lower bound for the marginal:
\begin{align}
\label{eq:x}
\begin{split}
    \mathcal{L}_{x} & = \mathbb{E}_{q(y|z, x)} \mathbb{E}_{q(z|x)} \log p(x|z,y) \\
    & \,\, - \text{KL} [q(z|x) || p(z)] 
\end{split}
\end{align}
where the first term is the auto-encoder reconstruction of $x$ with a cascaded NLU-NLG path.
The second term is the KL divergence which regularizes the approximated posterior distribution.
Detailed derivations can be found in Appendix.

When computing the reconstruction term of $x$, it requires us to first run through the NLU model to obtain the prediction on $y$, from which we run through NLG to reconstruct $x$.
The full information flow is ($x \hspace{-1mm} \rightarrow \hspace{-1mm} z \hspace{-1mm} \rightarrow \hspace{-1mm} y \hspace{-1mm} \rightarrow \hspace{-1mm} z \hspace{-1mm} \rightarrow \hspace{-1mm} x $).\footnote{This information flow requires us to sample both $z$ and $y$ in reconstructing $x$. Since $y$ is a discrete sequence, we use REINFORCE \cite{williams1992simple} to pass the gradient from NLG to NLU in the cascaded NLU-NLG path.}
Connections can be drawn with recent work which uses back-translation to augment training data for machine translation \cite{sennrich2016improving,he2016dual}. Unlike back-translation, the presence of latent variable in our model requires us to sample $z$ along the NLU-NLG path. The introduced stochasticity allows the model to explore a larger area of the data manifold.

The above describes the objectives when we have unlabelled $x$.
We can derive a similar objective for leveraging unlabelled $y$:
\begin{align}
\label{eq:y}
\begin{split}
   \mathcal{L}_{y} & = \mathbb{E}_{q(x|z, y)} \mathbb{E}_{q(z|y)} \log p(y|z,x) \\
    & \,\, - \text{KL} [q(z|y) || p(z)] 
\end{split}
\end{align}
where the first term is the auto-encoder reconstruction of $y$ with a cascaded NLG-NLU path. The full information flow here is ($y \hspace{-1mm} \rightarrow \hspace{-1mm} z \hspace{-1mm} \rightarrow \hspace{-1mm} x \hspace{-1mm} \rightarrow \hspace{-1mm} z \hspace{-1mm} \rightarrow \hspace{-1mm} y$).

\subsection{Choice of Prior \label{choice of p}}
The objectives described in \ref{sec:p_xy} and \ref{sec:p_x} require us to match an approximated posterior (either $q(z|x)$ or $q(z|y)$) to a prior $p(z)$ that reflects our belief.
A common choice of $p(z)$ in the research literature is the Normal distribution \cite{kingma2013auto}.
However, it should be noted that even if we match both $q(z|x)$ and $q(z|y)$ to the same prior, it does not guarantee that the two inferred posteriors are close to each other; this is a desired property of the shared latent space.

To better address the property, we propose a novel prior choice: when the posterior is inferred from $x$ (i.e., $q(z|x)$), we choose the parameterised distribution $q(z|y)$ as our prior belief of $p(z)$. 
Similarly, when the posterior is inferred from $y$ (i.e., $q(z|y)$), we have the freedom of defining $p(z)$ to be $q(z|x)$. This approach directly pulls $q(z|x)$ and $q(z|y)$ closer to ensure a shared latent space. 

Finally, note that it is straightforward to compute both $q(z|x)$ and $q(z|y)$ when we have parallel $x$ and $y$.
However when we have the access to unlabelled data, as described in Section \ref{sec:p_x}, we can only use the pseudo $x$-$y$ pairs that are generated by our NLU or NLG model, such that we can match an inferred posterior to a pre-defined prior reflecting our belief of the shared latent space.

\subsection{Training Summary \label{opt summary}}
In general, JUG subsumes the following three training scenarios which we will experiment with.

When we have fully labelled $x$ and $y$, the JUG jointly optimises NLU and NLG in a supervised fashion with the objective as follows:
\begin{equation}
\label{combine}
    \mathcal{L}_{basic} = \sum_{(x,y) \sim (X,Y) }( \mathcal{L}_{x, y} + \mathcal{L}_{y, x})
\end{equation}
where $(X, Y)$ denotes the set of labelled examples.

Additionally in the fully supervised setting, JUG can be trained to optimise both NLU, NLG and auto-encoding paths. This corresponds to the following objective:
\begin{equation}
\label{marginal}
    \mathcal{L}_{marginal} = \mathcal{L}_{basic} + \sum_{(x,y) \sim (X,Y) }(\mathcal{L}_{x} + \mathcal{L}_{y})
\end{equation}

Furthermore, when we have additional unlabelled $x$ or $y$, we optimise a semi-supervised JUG objective as follows:
\begin{equation}
\label{semi}
    \mathcal{L}_{semi} = \mathcal{L}_{basic} + \sum_{x \sim X}  \mathcal{L}_{x} + \sum_{y \sim Y}  \mathcal{L}_{y}
\end{equation}
where $X$ denotes the set of utterances and $Y$ denotes the set of formal representations.

\section{Experiments}
We experiment on two dialogue datasets with different formal representations to test the generality of our model.
The first dataset is \textbf{E2E} \cite{novikova-etal-2017-e2e}, which contains utterances annotated with flat slot-value pairs as their semantic representations.
The second dataset is the recent \textbf{weather} dataset \cite{balakrishnan2019constrained}, where both utterances and semantics are represented in tree structures. 
Examples of the two datasets are provided in tables \ref{tab:example_e2e} and \ref{tab:example_weather}.

\begin{table}[!t]
\centering
\resizebox{0.9\columnwidth}{!}{%
\begin{tabu}{c}
\tabucline [1pt]{1}
\textbf{Natural Language} \\
\textit{\begin{tabular}[c]{@{}c@{}}"sousa offers british food in the low price range. \\ it is family friendly with a 3 out of 5 star rating.\\ you can find it near the sunshine vegetarian cafe."\end{tabular}} \\ \hline
\textbf{Semantic Representation} \\
\begin{tabular}[c]{@{}c@{}}restaurant\_name=sousa, food=english,\\ price\_range=cheap, customer\_rating=average,\\ family\_friendly=yes, near=sunshine vegetarian cafe\end{tabular} \\
\tabucline [1pt]{1}
\end{tabu}%
}
\caption{An example in E2E dataset.}
\label{tab:example_e2e}
\end{table}

\begin{table}[!t]
\centering
\resizebox{\columnwidth}{!}{%
\begin{tabu}{c}
\tabucline [1pt]{1}
\textbf{Natural Language (original)} \\
\begin{tabular}[c]{@{}c@{}}"{[}\_\_DG\_YES\_\_ \textit{Yes} {]} , {[}\_\_DG\_INFORM\_\_\\ {[}\_\_ARG\_DATE\_TIME\_\_  {[}\_\_ARG\_COLLOQUIAL\_\_ \textit{today's} {]} {]}\\ \textit{forecast is}  {[}\_\_ARG\_CLOUD\_COVERAGE\_\_ \textit{mostly cloudy} {]}\\ \textit{with} {[}\_\_ARG\_CONDITION\_\_ \textit{light rain showers} {]} {]} ."\end{tabular} \\ \hline
\textbf{Natural Language (processed by removing tree annotations)} \\
\textit{"Yes, today's forecast is mostly cloudy with light rain showers."} \\ \hline
\textbf{Semantic Representation} \\
\begin{tabular}[c]{@{}c@{}}{[}\_\_DG\_YES\_\_ {[}\_\_ARG\_TASK\_\_ get\_weather\_attribute {]} {]} \\ {[}\_\_DG\_INFORM\_\_ {[}\_\_ARG\_TASK\_\_ get\_forecast {]}\\ {[}\_\_ARG\_CONDITION\_\_ light rain showers {]}\\ {[}\_\_ARG\_CLOUD\_COVERAGE\_\_ mostly cloudy {]}\\ {[}\_\_ARG\_DATE\_TIME\_\_ {[}\_\_ARG\_COLLOQUIAL\_\_ today's {]} {]} {]}\end{tabular} \\
\tabucline [1pt]{1}
\end{tabu}%
}
\caption{An example in weather dataset. The natural language in original dataset (first row) is used for training to have a fair comparison with existing methods. The processed utterances (second row) is used in our semi-supervised setting.}
\label{tab:example_weather}
\end{table}

\begin{table}[t]
\centering
\resizebox{0.65\columnwidth}{!}{%
\begin{tabu}{llll}
\tabucline [1pt]{1-4}
Dataset & Train & Valid & Test \\ \hline
E2E & 42061 & 4672 & 4693 \\
Weather & 25390 & 3078 & 3121 \\
\tabucline [1pt]{1-4}
\end{tabu}%
}
\caption{Number of examples in two datasets}
\label{tab:statistics}
\end{table}

\begin{table}[t]
\centering
\resizebox{\columnwidth}{!}{%
\begin{tabu}{ll}
\tabucline [1pt]{1-2}
\textbf{E2E NLU} & \textbf{F1} \\
Dual supervised learning \cite{su2019dual} & 0.7232 \\
$\textup{JUG}_{\textup{basic}}$ & \textbf{0.7337} \\ \hline \hline
\textbf{E2E NLG} & \textbf{BLEU} \\
TGEN \cite{duvsek2016sequence} & 0.6593 \\
SLUG \cite{juraska2018deep} & 0.6619 \\
Dual supervised learning \cite{su2019dual} & 0.5716 \\
$\textup{JUG}_{\textup{basic}}$ & \textbf{0.6855} \\ \hline \hline
\textbf{Weather NLG} & \textbf{BLEU} \\
S2S-CONSTR \cite{balakrishnan2019constrained} & 0.7660 \\
$\textup{JUG}_{\textup{basic}}$ & \textbf{0.7768} \\
\tabucline [1pt]{1-2}
\end{tabu}%
}
\caption{Comparison with previous systems on two datasets. Note that there is no previous system trained for NLU in weather dataset.}
\label{tab:sota}
\end{table}

\begin{table*}[t]
\centering
\resizebox{0.9\textwidth}{!}{%
\begin{tabu}{lccccc}
\tabucline [1pt]{1-7}
Model / Data & 5\% & 10\% & 25\% & 50\% & 100\% \\ \hline
Decoupled & 52.77 (0.874) & 62.32 (0.902) & 69.37 (0.924) & 73.68 (0.935) & 76.12 (0.942) \\
$\textup{Augmentation}^{*}$ & 54.71 (0.878) & 62.54 (0.902) & 68.91 (0.922) & 73.84 (0.935) & - \\
$\textup{JUG}_{\textup{basic}}$ & 60.30 (0.902) & 67.08 (0.918) & 72.49 (0.932) & 74.74 (0.937) & 78.05 (0.945) \\
$\textup{JUG}_{\textup{marginal}}$ & 62.96 (0.907) & 68.43 (0.920) & 73.35 (0.933) & \textbf{75.74 (0.939)} & \textbf{78.93 (0.948)} \\
$\textup{JUG}^{*}_{\textup{semi}}$ & \textbf{68.09 (0.921)} & \textbf{70.33 (0.925)} & \textbf{73.79 (0.935)} & 75.46 (0.939) & - \\
\tabucline [1pt]{1-7}
\end{tabu}%
}
\caption{NLU results on E2E dataset. Joint accuracy (\%) and F1 score (in bracket) are both reported with varying percentage of labelled training data. Models using unlabelled data are marked with *.}
\label{tab:nlu_e2e}
\end{table*}


\begin{table*}[t]
\LARGE
\centering
\resizebox{0.9\textwidth}{!}{%
\begin{tabu}{lccccc}
\tabucline [1pt]{1-7}
Model / Data & 5\% & 10\% & 25\% & 50\% & 100\% \\ \hline
Decoupled & 0.693 (83.47) & 0.723 (87.33) & 0.784 (92.52) & 0.793 (94.91) & 0.813 (96.98) \\
$\textup{Augmentation}^{*}$ & 0.747 (84.79) & 0.770 (90.13) & 0.806 (94.06) & 0.815 (96.04) & - \\
$\textup{JUG}_{\textup{basic}}$ & 0.685 (84.20) & 0.734 (88.68) & 0.769 (93.83) & 0.788 (95.11) & 0.810 (95.07) \\
$\textup{JUG}_{\textup{marginal}}$ & 0.724 (85.57) & 0.775 (93.59) & 0.803 (94.99) & 0.817 (98.67) & \textbf{0.830 (99.11)} \\
$\textup{JUG}^{*}_{\textup{semi}}$ & \textbf{0.814 (90.47)} & \textbf{0.792 (94.76)} & \textbf{0.819 (95.59)} & \textbf{0.827 (98.42)} & - \\ 
\tabucline [1pt]{1-7}
\end{tabu}%
}
\caption{NLG results on E2E dataset. BLEU and semantic accuracy (\%) (in bracket) are both reported with varying percentage of labelled training data. Models using unlabelled data are marked with *.}
\label{tab:nlg_e2e}
\end{table*}

\subsection{Training Scenarios}
\label{training_scenarios}
We primarily evaluated our models on the raw splits of the original datasets, which enables us to fairly compare fully-supervised JUG with existing work on both NLU and NLG.
\footnote{Following \citet{balakrishnan2019constrained}, the evaluation code \href{https://github.com/tuetschek/e2e-metrics}{https://github.com/tuetschek/e2e-metrics} provided by the E2E organizers is used here for calculating BLEU in NLG.}
Statistics of the two datasets can be found in Table \ref{tab:statistics}.

In addition, we set up an experiment to evaluate semi-supervised JUG with a varying amount of labelled training data (5\%, 10\%, 25\%, 50\%, 100\%, with the rest being unlabelled).
Note that the original E2E test set is designed on purpose with unseen slot-values in the test set to make it difficult \cite{duvsek2018findings, duvsek2020evaluating}; we remove the distribution bias by randomly re-splitting the E2E dataset. 
On the contrary, utterances in the weather dataset contains extra tree-structure annotations which make the NLU task a toy problem. We therefore remove these annotations to make NLU more realistic, as shown in the second row of Table \ref{tab:example_weather}.

As described in Section \ref{opt summary}, we can optimise our proposed JUG model in various ways. We investigate the following approaches:

$\textbf{\texttt{JUG}}_{\textbf{\texttt{basic}}}$: this model jointly optimises NLU and NLG with the objective in Equation \ref{combine}. This uses labelled data only.

$\textbf{\texttt{JUG}}_{\textbf{\texttt{marginal}}}$: jointly optimises NLU, NLG and auto-encoders with only labelled data, per Equation \ref{marginal}. 

$\textbf{\texttt{JUG}}_{\textbf{\texttt{semi}}}$: jointly optimises NLU and NLG  with labelled data and auto-encoders with unlabelled data, per Equation \ref{semi}.

\subsection{Baseline Systems}
We compare our proposed model with some existing methods as shown in Table \ref{tab:sota} and two designed baselines as follows:

\textbf{\texttt{Decoupled}}: The NLU and NLG models are trained separately by supervised learning. Both of the individual models have the same encoder-decoder structure as JUG. However, the main difference is that there is no shared latent variable between the two individual NLU and NLG models.

\textbf{\texttt{Augmentation}}: We pre-train \texttt{Decoupled} models to generate pseudo label from the unlabelled corpus \cite{lee2013pseudo} in a setup similar to back-translation \cite{sennrich2016improving}. The pseudo data and labelled data are then used together to fine-tune the pre-trained models.

Among all systems in our experiments, the number of units in LSTM encoder/decoder are set to \{150, 300\} and the dimension of latent space is 150. The optimiser Adam \cite{kingma2014adam} is used with learning rate 1e-3.
Batch size is set to \{32, 64\}. All the models are fully trained and the best model is picked by the average of NLU and NLG results on validation set during training.
The source code can be found at \url{https://github.com/andy194673/Joint-NLU-NLG}.

\begin{table}[t]
\huge
\centering
\resizebox{\columnwidth}{!}{%
\begin{tabu}{lccccc}
\tabucline [1pt]{1-7}
Model / Data & 5\% & 10\% & 25\% & 50\% & 100\% \\ \hline
Decoupled & 73.46 & 80.85 & 86.00 & 88.45 & 90.68 \\
$\textup{Augmentation}^{*}$ & 74.77 & 79.84 & 86.24 & 88.69 & - \\
$\textup{JUG}_{\textup{basic}}$ & 73.62 & 80.13 & 86.15 & 87.94 & 90.55 \\
$\textup{JUG}_{\textup{marginal}}$ & 74.61 & 81.14 & 86.83 & 89.06 & \textbf{91.28} \\
$\textup{JUG}^{*}_{\textup{semi}}$ & \textbf{79.19} & \textbf{83.22} & \textbf{87.46} & \textbf{89.17} & - \\
\tabucline [1pt]{1-7}
\end{tabu}%
}
\caption{NLU results with exact match accuracy (\%) on weather dataset.}
\label{tab:nlu_weather}
\end{table}

\subsection{Main Results}
\label{sec:main-results}
We start by comparing the $\texttt{JUG}_{\texttt{basic}}$ performance with existing work following the original split of the datasets. The results are shown in Table \ref{tab:sota}.  
On E2E dataset, we follow previous work to use F1 of slot-values as the measurement for NLU, and BLEU-4 for NLG.
For weather dataset, there is only published results for NLG.
It can be observed that the $\texttt{JUG}_{\texttt{basic}}$ model outperforms the previous state-of-the-art NLU and NLG systems on the E2E dataset, and also for NLG on the weather dataset. The results prove the effectiveness of introducing the shared latent variable $z$ for jointly training NLU and NLG.
We will further study the impact of the shared $z$ in Section \ref{sharingz}.

\begin{table}[t]
\huge
\centering
\resizebox{\columnwidth}{!}{%
\begin{tabu}{lccccc}
\tabucline [1pt]{1-7}
Model / Data & 5\% & 10\% & 25\% & 50\% & 100\% \\ \hline
Decoupled & 0.632 & 0.667 & 0.703 & 0.719 & 0.725 \\
$\textup{Augmentation}^{*}$ & 0.635 & 0.677 & 0.703 & 0.727 & - \\
$\textup{JUG}_{\textup{basic}}$& 0.634 & 0.673 & 0.701 & 0.720 & \textbf{0.726} \\
$\textup{JUG}_{\textup{marginal}}$ & 0.627 & 0.671 & 0.711 & 0.721 & 0.722 \\
$\textup{JUG}^{*}_{\textup{semi}}$& \textbf{0.670} & \textbf{0.701} & \textbf{0.725} & \textbf{0.733} & - \\ 
\tabucline [1pt]{1-7}
\end{tabu}%
}
\caption{NLG results with BLEU on weather dataset.}
\label{tab:nlg_weather}
\end{table}

We also evaluated the three training scenarios of JUG in the semi-supervised setting, with different proportion of labelled and unlabelled data. 
The results for E2E is presented in Table \ref{tab:nlu_e2e} and  \ref{tab:nlg_e2e}.
We computed both F1 score and joint accuracy \cite{mrkvsic2017neural} of slot-values as a more solid NLU measurement. Joint accuracy is defined as the proportion of test examples whose slot-value pairs are all correctly predicted. For NLG, both BLEU-4 and semantic accuracy are computed. Semantic accuracy measures the proportion of correctly generated slot values in the produced utterances. From the results, we observed that \texttt{Decoupled} can be improved with techniques of generating pseudo data (\texttt{Augmentation}), which forms a stronger baseline.
However, all our model variants perform better than the baselines on both NLU and NLG.
When using only labelled data, our model $\texttt{JUG}_{\texttt{marginal}}$ can surpass \texttt{Decoupled} across all the four measurements.
The gains mainly come from the fact that the model uses auto-encoding objectives to help learn a shared semantic space. Compared to \texttt{Augmentation}, $\texttt{JUG}_{\texttt{marginal}}$ also has a `built-in mechanism' to bootstrap pseudo data on the fly of training (see Section \ref{opt summary}). 
When adding extra unlabelled data, our model $\texttt{JUG}_{\texttt{semi}}$ gets further performance boosts and outperforms all baselines by a significant margin.

With the varying proportion of unlabelled data in the training set, we see that unlabelled data is helpful in almost all cases. Moreover, the performance gain is the more significant when the labelled data is less. This indicates that the proposed model is especially helpful for low resource setups when there is a limited amount of labelled training examples but more available unlabelled ones.

The results for weather dataset are presented in Table \ref{tab:nlu_weather} and \ref{tab:nlg_weather}.
In this dataset, NLU is more like a semantic parsing task \cite{berant-etal-2013-semantic} and we use exact match accuracy as its measurement. Meanwhile, NLG is measured by BLEU. 
The results reveal a very similar trend to that in E2E.
The generated examples can be found in Appendix.

\begin{figure}[t]
  \centering
  \includegraphics[width=\columnwidth]{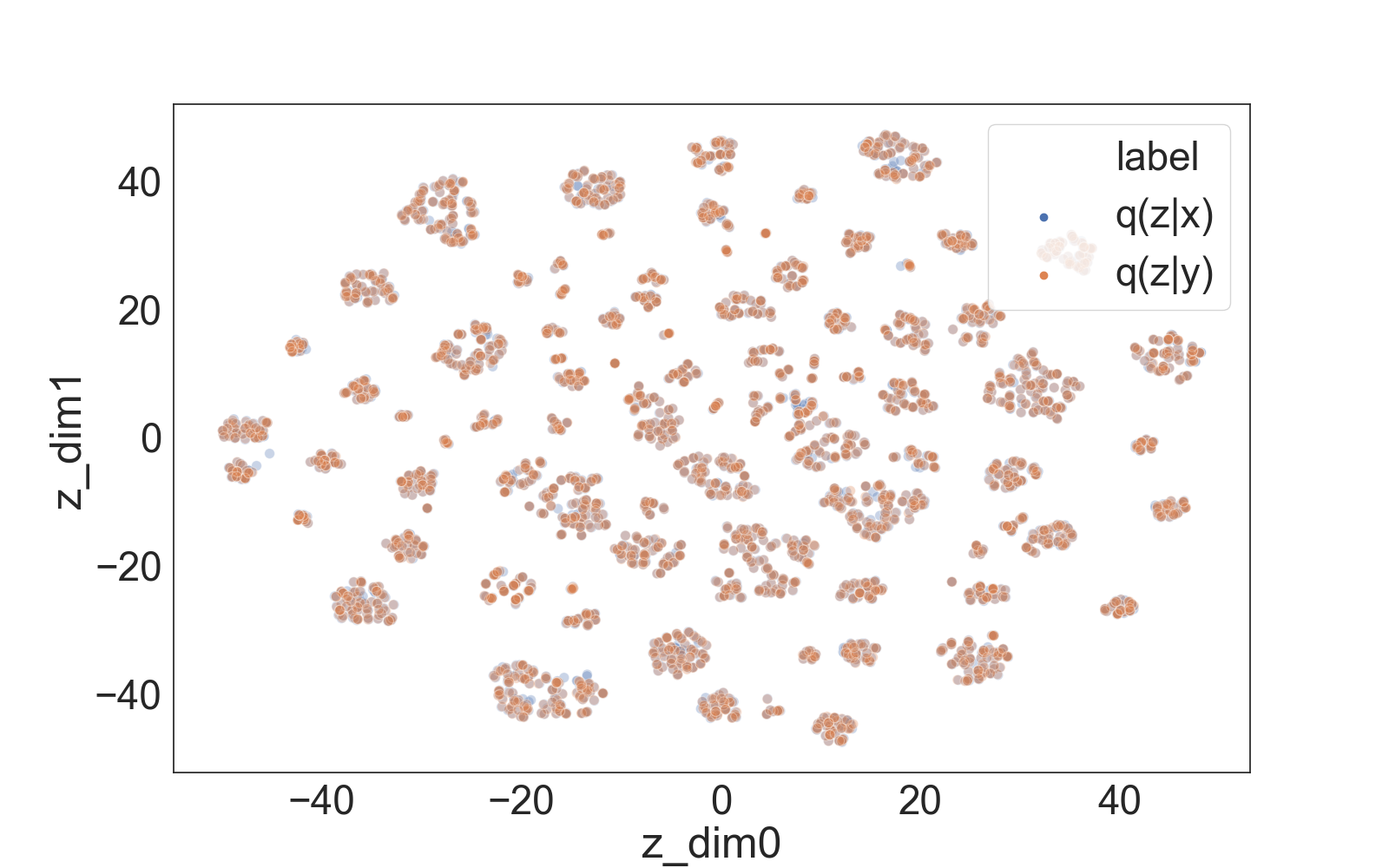}
  \caption{Visualisation of latent variable $z$. Given a pair of $x$ and $y$, $z$ can be sampled from the posterior $q(z|x)$ or $q(z|y)$, denoted by blue and orange dots respectively.}
  \label{fig:z_space}
  \vspace{-0.3em}

\end{figure}

\subsection{Analysis}
In this section we further analyse  the impact of the shared latent variable and also the impact of utilising unlabelled data.

\subsubsection{Visualisation of Latent Space}
As mentioned in Section \ref{sec:nlu-nlg}, the latent variable $z$ can be sampled from either posterior approximation $q(z|x)$ or $q(z|y)$. We inspect the latent space in Figure \ref{fig:z_space} to find out how well the model learns intent sharing. 
We plot $z$ with the E2E dataset on 2-dimentional space using t-SNE projection \cite{maaten2008visualizing}.

We observe two interesting properties. First, for each data point ($x$, $y$), the $z$ values sampled from $q(z|x)$ and $q(z|y)$ are close to each other. This reveals that the meanings of $x$ and $y$ are tied in the latent space.
Second, there exists distinct clusters in the space of $z$. By further inspecting the actual examples within each cluster, we found that a cluster represents a similar meaning composition. For instance, the cluster centered at (-20, -40) contains \{\texttt{name}, \texttt{foodtype}, \texttt{price}, \texttt{rating}, \texttt{area}, \texttt{near}\}, while the cluster centered at (45, 10) contains \{\texttt{name}, \texttt{eattype}, \texttt{foodtype}, \texttt{price}\}. This indicates that the shared latent serves as conclusive global feature representations for NLU and NLG.

\subsubsection{Impact of the Latent Variable \label{sharingz}}
One novelty of our model is the introduction of shared latent variable $z$ for natural language $x$ and formal representations $y$.
A common problem in neural variational models is that when coupling a powerful autogressive decoder, the decoder tends to learn to ignore $z$ and solely rely on itself to generate the data \cite{bowman2016generating,chen2017variational,goyal2017z}.
In order to examine to what extent does our model actually rely on the shared variable in both NLU and NLG, we seek for an empirical answer by comparing the $\texttt{JUG}_{\texttt{basic}}$ model with a model variant which uses a random value of $z$ sampled from a normal distribution $N(\mathbf{0},\mathbf{1})$ during testing. From Table \ref{tab:z_test},
we can observe that there exists a large performance drop if $z$ is assigned with random values. This suggests that JUG indeed relies greatly on the shared variable to produce good-quality $x$ or $y$.

\begin{table}[]
\centering
\resizebox{0.85\columnwidth}{!}{%
\begin{tabu}{lll}
\tabucline [1pt]{1-3}
Model & NLU & NLG \\ \hline
$\textup{JUG}_{\textup{basic}}$ & 90.55 & 0.726 \\
$\textup{JUG}_{\textup{basic}}$ (feed random z) & 38.13 & 0.482 \\
\tabucline [1pt]{1-3}
\end{tabu}%
}
\caption{A comparative study to evaluate the contribution of the learned latent variable $z$ in NLU/NLG decoding. Models are trained on the whole weather dataset.}
\label{tab:z_test}
\end{table}

\begin{table}[t]
\centering
\resizebox{0.85\columnwidth}{!}{%
\begin{tabu}{lccccc}
\tabucline [1pt]{1-6}
\multirow{2}{*}{Method} & \multicolumn{3}{c}{NLU} & \multicolumn{2}{c}{NLG} \\
 & Mi & Re & Wr & Mi & Wr \\ \hline
Decoupled & 714 & 256 & 2382 & 5714 & 2317 \\
$\textup{JUG}_{\textup{basic}}$ & \textbf{594} & \textbf{169} & \textbf{1884} & \textbf{4871} & \textbf{2102} \\
\tabucline [1pt]{1-6}
\end{tabu}%
}
\caption{Error analysis on E2E dataset. Numbers of missing (Mi), redundant (Re) and wrong (Wr) predictions on slot-value pairs are reported for NLU; numbers of missing or wrong generated slot values are listed for NLG. Lower number indicates the better results. Both models are trained on 5\% of the training data.}
\label{tab:error_analysis}
    \vspace{-0.3em}
\end{table}

We further analyse the various sources of errors to understand the cases which $z$ helps to improve.
On E2E dataset, wrong prediction in NLU comes from either predicting \texttt{not\_mention} label for certain slots in ground truth semantics; predicting arbitrary values on slots not present in the ground truth semantics; or predicting wrong values comparing to ground truth. Three types of error are referred to Missing (Mi), Redundant (Re) and Wrong (Wr) in Table \ref{tab:error_analysis}. For NLG, semantic errors can be either missing or generating wrong slot values in the given semantics \cite{wen2015semantically}.
Our model makes fewer mistakes in all these error sources comparing to the baseline \texttt{Decoupled}. We believe this is because the clustering property learned in the latent space provides better feature representations at a global scale, eventually benefiting NLU and NLG.

\subsubsection{Impact of Unlabelled Data Source}
In Section \ref{sec:main-results}, we found that the performance of our model can be further enhanced by leveraging unlabelled data. As we used both unlabelled utterances and unlabelled semantic representations together, it is unclear if both contributed to the performance gain. 
To answer this question, we start with the $\texttt{JUG}_{\texttt{basic}}$ model, and experimented with adding unlabelled data from 1) only unlabelled utterances $x$; 2) only semantic representations $y$; 3) both $x$ and $y$. As shown in Table \ref{tab:source}, when adding any uni-sourced unlabelled data ($x$ or $y$), the model is able to improve to a certain extent. However, the performance can be maximised when both data sources are utilised. This strengthens the argument that our model can leverage bi-sourced unlabelled data more effectively via latent space sharing to improve NLU and NLG at the same time.

\begin{table}[t]
\LARGE
\centering
\resizebox{0.95\columnwidth}{!}{%
\begin{tabu}{lcccc}
\tabucline [1pt]{1-5}
 & \multicolumn{2}{c}{E2E} & \multicolumn{2}{c}{Weather} \\
Method & NLU & NLG & NLU & NLG \\
\tabucline [1pt]{1-5}
$\textup{JUG}_{\textup{basic}}$ & 60.30 & 0.685 & 73.62 & 0.634 \\ \hline
\,\,\, +unlabelled $x$ & 62.89 & 0.765 & 74.97 & 0.654 \\
\,\,\, +unlabelled $y$ & 59.55 & \textbf{0.815} & 76.98 & 0.621 \\
\,\,\, +unlabelled $x$ and $y$ & \textbf{68.09} & 0.814 & \textbf{79.19} & \textbf{0.670} \\
\tabucline [1pt]{1-5}
\end{tabu}%
}
\caption{Comparison on sources of unlabelled data for semi-supervised learning using only utterances ($x$), only semantic representations ($y$) or both ($x$ and $y$). $\texttt{JUG}_{\texttt{basic}}$ model is trained on 5\% of training data.}
\label{tab:source}
    \vspace{-0.3em}
\end{table}

\section{Related Work}
Natural Language Understanding (NLU) refers to the general task of mapping natural language to formal representations.
One line of research in the dialogue community aims at detecting slot-value pairs expressed in user utterances as a classification problem \cite{henderson2012discriminative,sun2014sjtu,mrkvsic2017neural,vodolan2017hybrid}.
Another line of work focuses on converting single-turn user utterances to more structured meaning representations as a semantic parsing task \cite{zettlemoyer2005learning,jia2016data,dong2018coarse,damonte2019practical}.

In comparison, Natural Language Generation (NLG) is scoped as the task of generating natural utterances from their formal representations. This is traditionally handled with a pipelined approach \cite{reiter1997building} with content planning and surface realisation \cite{walker2001spot,stent2004trainable}. More recently, NLG has been formulated as an end-to-end learning problem where text strings are generated with recurrent neural networks conditioning on the formal representation \cite{wen2015semantically,duvsek2016sequence,duvsek2020evaluating,balakrishnan2019constrained, tseng2019tree}.

There has been very recent work which does NLU and NLG jointly.
Both \citet{ye2019jointly} and \citet{cao2019semantic} explore the duality of semantic parsing and NLG. The former optimises two sequence-to-sequence models using dual information maximisation, while the latter introduces a dual learning framework for semantic parsing.
\citet{su2019dual} proposes a learning framework for dual supervised learning \cite{xia2017dual} where both NLU and NLG models are optimised towards a joint objective.
Their method brings benefits with annotated data in supervised learning, but does not allow semi-supervised learning with unlabelled data.
In contrast to their work, we propose a generative model which couples NLU and NLG with a shared latent variable.
We focus on exploring a coupled representation space between natural language and corresponding semantic annotations.
As proved in experiments, the information sharing helps our model to leverage unlabelled data for semi-supervised learning, which eventually benefits both NLU and NLG.

\section{Conclusion}
We proposed a generative model which couples natural language and formal representations via a shared latent variable.
Since the two space is coupled,
we gain the luxury of exploiting each unpaired data source and transfer the acquired knowledge to the shared meaning space. This eventually benefits both NLU and NLG, especially in a low-resource scenario. The proposed model is also suitable for other translation tasks between two modalities.

As a final remark, natural language is richer and more informal. NLU needs to handle ambiguous or erroneous user inputs. However, formal representations utilised by an NLG system are more precisely-defined. In future, we aim to refine our generative model to better emphasise this difference of the two tasks.


\section*{Acknowledgments}
Bo-Hsiang Tseng is supported by Cambridge Trust and the Ministry of Education, Taiwan. This work has been performed using resources provided by the Cambridge Tier-2 system operated by the University of Cambridge Research Computing Service (http://www.hpc.cam.ac.uk) funded by EPSRC Tier-2 capital grant EP/P020259/1..

\bibliography{acl2020}
\bibliographystyle{acl_natbib}

\clearpage

\appendix
\section{Appendices}
\subsection{Derivation of Lower Bounds}
We derive the lower bounds for $\log p(x,y)$ as follows:
\begin{align}
\begin{split}
    & \log p(x,y) = \log \int_{z} p(x,y,z) \\
    & \quad\quad\quad = \log \int_z \frac{p(x,y,z)q(z|x)}{q(z|x)} \\
    & \quad\quad\quad = \log \int_z \frac{p(x|z, y)p(y|z, x)p(z)q(z|x)}{q(z|x)} \\
    & \quad\quad\quad = \log \mathbb{E}_{q(z|x)} \frac{p(x|z, y)p(y|z, x)p(z)}{q(z|x)} \\
    & \quad\quad\quad \geq \mathbb{E}_{q(z|x)} \log
    \frac{p(x|z, y)p(y|z, x)p(z)}{q(z|x)} \\
    & \quad\quad\quad = \mathbb{E}_{q(z|x)} [\log p(x|z, y) + \log p(y|z, x)] \\
    & \quad\quad\quad\quad - \text{KL} [q(z|x) || p(z)]
\end{split}
\end{align}
where $q(z|x)$ represents an approximated posterior. This derivation gives us the Equation 6 in the paper. Similarly we can derive an alternative lower bound in Equation 7 by introducing $q(z|y)$ instead of $q(z|x)$.

For marginal log-likelihood $\log p(x)$ or  $\log p(y)$, its lower bound is derived as follows:
\begin{align}
\begin{split}
    & \log p(x) = \log \int_y \int_{z} p(x,y,z) \\
    & \quad = \log \int_{y} \int_{z} \frac{p(x|z,y) p(y) p(z) q(z|x) q(y|z,x)}{q(z|x) q(y|z, x)} \\
    & \quad = \log \mathbb{E}_{q(y|z, x)} \mathbb{E}_{q(z|x)} \frac{p(x|z,y) p(y) p(z)}{q(z|x) q(y|z, x) } \\
    & \quad \geq \mathbb{E}_{q(y|z, x)} \mathbb{E}_{q(z|x)} \log \frac{p(x|z,y) p(y) p(z)}{q(z|x) q(y|z, x)} \\
    & \quad = \mathbb{E}_{q(y|z, x)} \mathbb{E}_{q(z|x)} \log p(x|z,y) \\
    & \quad\quad - \text{KL} [q(z|x) || p(z)] - \text{KL} [q(y|x, z) || p(y)]
\end{split}
\end{align}
Note that the resulting lower bound consists of three terms: a reconstruction of $x$, a KL divergence which regularises the space of $z$, and also a KL divergence which regularises the space of $y$. 
We have dropped the last term in our optimisation objective in Equation 9, since we do not impose any prior assumption on the output space of the NLU model. 

Analogously we can derive the lower bound for $\log p(y)$.
We also do not impose any prior assumption on the output space of the NLG model, which leads us to Equation 10.

\onecolumn
\subsection{Generated Examples}
\begin{table*}[!ht]
\centering
\resizebox{\textwidth}{!}{%
\begin{tabu}{l}
\tabucline [2pt]{1}
Reference of example \\
x: \textit{"for those prepared to pay over $\pounds$30 , giraffe is a restaurant located near the six bells ."} \\
y: \{name=giraffe, eat\_type=restaurant, price\_range=more than $\pounds$30, near=the six bells\} \\ \tabucline [1pt]{1}
Prediction by Decoupled model \\
x: \textit{"near the six bells , there is a restaurant called giraffe that is children friendly ."} \textcolor{bostonuniversityred}{\textsf{(miss price\_range)}}  \\
y: \{name=travellers rest beefeater, price\_range=more than $\pounds$30, near=the six bells\} \textcolor{bostonuniversityred}{\textsf{(wrong name, miss eat\_type)}}  \\ \tabucline [1pt]{1}
Prediction by JUG$_{\textup{semi}}$ model \\
x: \textit{"giraffe is a restaurant near the six bells with a price range of more than $\pounds$30 ."} \textcolor{cadmiumgreen}{\textsf{(semantically correct)}} \\
y: \{name=giraffe, eat\_type=restaurant, price\_range=more than $\pounds$30, near=the six bells\} \textcolor{cadmiumgreen}{\textsf{(exact match)}} \\
\tabucline [2pt]{1}
\end{tabu}%
}
\caption{An example of E2E dataset and predictions generated by the baseline model $\texttt{Decoupled}$ and the proposed model $\texttt{JUG}_{\texttt{semi}}$. $x$ and $y$ denotes natural language and the corresponding semantic representation. Errors are highlighted following predictions.}
\end{table*}

\begin{table*}[!ht]
\centering
\resizebox{\textwidth}{!}{%
\begin{tabu}{l}
\tabucline [2pt]{1}
Reference of example \\ \hline

x: \textit{"it's going to be} \_\_arg\_temp\_\_ \textit{and} \_\_arg\_cloud\_coverage\_\_ \\ \_\_arg\_colloquial\_\_ \textit{between} \_\_arg\_start\_time\_\_ \textit{and} \_\_arg\_end\_time\_\_\textit{"} \\ \hline

\begin{tabular}[c]{@{}l@{}}y: {[}\_\_dg\_inform\_\_ {[}\_\_arg\_task\_\_ get\_forecast {]} {[}\_\_arg\_temp\_\_ \_\_arg\_temp\_\_ {]} \\ {[}\_\_arg\_cloud\_coverage\_\_ \_\_arg\_cloud\_coverage\_\_ {]}\\ {[}\_\_arg\_ date\_time\_range\_\_ {[}\_\_arg\_start\_time\_\_ \_\_arg\_start\_time\_\_ {]} \\ {[}\_\_arg\_end\_time\_\_ \_\_arg\_end\_time\_\_ {]} {[}\_\_arg\_colloquial\_\_ \_\_arg\_colloquial\_\_ {]} {]} {]}\end{tabular} \\ \tabucline [1pt]{1}

Prediction by Decoupled model \\ \hline

x: \textit{"it will be} \_\_arg\_temp\_\_ degrees \textit{and} \_\_arg\_cloud\_coverage\_\_ \textit{from} \\ \_\_arg\_start\_time\_\_ \textit{to} \_\_arg\_end\_time\_\_" \\ \hline

\begin{tabular}[c]{@{}l@{}}y: {[}\_\_dg\_inform\_\_ {[}\_\_arg\_task\_\_ get\_forecast {]} {[}\_\_arg\_temp\_\_ \_\_arg\_temp\_\_ {]} {[}\_\_arg\_cloud\_coverage\_\_ \\ \_\_arg\_cloud\_coverage\_\_ {]} {[}\_\_arg\_ date\_time\_\_ {[}\_\_arg\_colloquial\_\_ \_\_arg\_colloquial\_\_ {]} {]} {]} \\ {[}\_\_dg\_inform\_\_ {[}\_\_arg\_task\_\_ get\_forecast {]} {[}\_\_arg\_temp\_\_ \_\_arg\_temp\_\_ {]} {[}\_\_arg\_cloud\_coverage\_\_\\ \_\_arg\_cloud\_coverage\_\_ {]} {[}\_\_arg\_date\_time\_range\_\_ {[}\_\_arg\_start\_time\_\_ \_\_arg\_start\_time\_\_ {]} \\ {[}\_\_arg\_end\_time\_\_ \_\_arg\_end\_time\_\_ {]} {]} {]} \textcolor{bostonuniversityred}{\textsf{(not match)}} \end{tabular} \\ \tabucline [1pt]{1}

Prediction by JUG\_semi model \\ \hline

x: \textit{"the temperature will be around} \_\_arg\_temp\_\_  \textit{degrees} \\ \_\_arg\_colloquial\_\_ \textit{between} \_\_arg\_start\_time\_\_ \textit{and} \_\_arg\_end\_time\_\_" \\ \hline

\begin{tabular}[c]{@{}l@{}}y: {[}\_\_dg\_inform\_\_ {[}\_\_arg\_task\_\_ get\_forecast {]} {[}\_\_arg\_temp\_\_ \_\_arg\_temp\_\_ {]} \\ {[}\_\_arg\_cloud\_coverage\_\_ \_\_arg\_cloud\_coverage\_\_ {]}\\ {[}\_\_arg\_ date\_time\_range\_\_ {[}\_\_arg\_start\_time\_\_ \_\_arg\_start\_time\_\_ {]} \\ {[}\_\_arg\_end\_time\_\_ \_\_arg\_end\_time\_\_ {]} {[}\_\_arg\_colloquial\_\_ \_\_arg\_colloquial\_\_ {]} {]} {]} \textcolor{cadmiumgreen}{\textsf{(exact match)}} \end{tabular} \\

\tabucline [2pt]{1}
\end{tabu}%
}
\caption{An example of weather dataset and predictions generated by the baseline model $\texttt{Decoupled}$ and the proposed model $\texttt{JUG}_{\texttt{semi}}$. $x$ and $y$ denotes natural language and the corresponding semantic representation. NLU result are highlighted following predictions.}
\end{table*}


\end{document}